\documentclass{article} %
\usepackage{iclr2026_conference,times}

\usepackage{amsmath,amsfonts,bm}

\def\eqref#1{equation~\ref{#1}}

\def\1{\bm{1}}

\DeclareMathAlphabet{\mathsfit}{\encodingdefault}{\sfdefault}{m}{sl}
\SetMathAlphabet{\mathsfit}{bold}{\encodingdefault}{\sfdefault}{bx}{n}

\newcommand{\eg}{\textit{e.g.}}

\definecolor{citecolor}{HTML}{2878b5}
\definecolor{linkcolor}{HTML}{c0392b}
\definecolor{boxcolor}{RGB}{194, 213, 247}
\definecolor{lightroyalblue}{HTML}{F6F8FD} 
\definecolor{boxcontentgray}{HTML}{F7F7F7}
\definecolor{boxtitlegray}{HTML}{CCCCCC}
\definecolor{boxbrown}{HTML}{D7CCC8}
\usepackage[hidelinks,breaklinks=true,colorlinks,bookmarks=false,citecolor=citecolor,linkcolor=linkcolor,urlcolor=linkcolor]{hyperref}

\usepackage{url}
\usepackage{enumitem}
\usepackage{booktabs}
\usepackage{multirow}
\usepackage{graphicx}
\usepackage[table,xcdraw]{xcolor}
\usepackage{wrapfig}
\usepackage{subcaption}
\usepackage[titletoc]{appendix}
\usepackage{titlesec}
\usepackage{titletoc}
\usepackage{setspace}
\usepackage{float}
\usepackage{algorithm}
\usepackage{comment}
\usepackage[noEnd=True]{algpseudocodex}
\usepackage{dashrule}
\usepackage[most]{tcolorbox}
\newtcolorbox{graybox}[1]{
  breakable,   
  fonttitle=\bfseries,
  enhanced,                        
  colback=boxcontentgray,        %
  colbacktitle=boxtitlegray,     %
  coltitle=black,                %
  colframe=black,                %
  coltext=black,                 %
  boxrule=0.5pt,
  arc=2mm,
  title=#1
}

\title{AdvChain: Adversarial Chain-of-Thought Tuning for Robust Safety Alignment of Large Reasoning Models}

\author{
\centerline{
Zihao Zhu\textsuperscript{\rm 1}\quad
Xinyu Wu\textsuperscript{\rm 1} \quad
Gehan Hu\textsuperscript{\rm 1} \quad
\textbf{Siwei Lyu}\textsuperscript{\rm 2} \quad
\textbf{Ke Xu}\textsuperscript{\rm 3} \quad
\textbf{Baoyuan Wu}\textsuperscript{\rm 1}\thanks{Corresponding Author}} \\
\centerline{\textsuperscript{1}The Chinese University of Hong Kong, Shenzhen} \\
\centerline{
\textsuperscript{2}State University of New York at Buffalo \quad
\textsuperscript{3}Huawei International, Singapore} \\
\centerline{\texttt{zihaozhu@link.cuhk.edu.cn}}
}

\iclrfinalcopy %
\begin{document}

\maketitle

\begin{abstract}
Large Reasoning Models (LRMs) have demonstrated remarkable capabilities in complex problem-solving through Chain-of-Thought (CoT) reasoning. However, the multi-step nature of CoT introduces new safety challenges that extend beyond conventional language model alignment. We identify a failure mode in current safety CoT tuning methods: the \textit{snowball effect}, where minor reasoning deviations progressively amplify throughout the thought process, leading to either harmful compliance or excessive refusal. This effect stems from models being trained to imitate perfect reasoning scripts without learning to self-correct. To address this limitation, we propose AdvChain, an alignment paradigm that teaches models dynamic self-correction through adversarial CoT tuning. Our method involves constructing a dataset containing Temptation-Correction and Hesitation-Correction samples, where models learn to recover from harmful reasoning drifts and unnecessary cautions. Extensive experiments show that AdvChain significantly enhances robustness against jailbreak attacks and CoT hijacking while substantially reducing over-refusal on benign prompts, achieving a superior safety-utility balance without compromising reasoning capabilities. Our work establishes a new direction for building more robust and reliable reasoning models.
\end{abstract}

\section{Introduction}

Large Reasoning Models (LRMs), which excel at complex problem-solving through explicit Chain-of-Thought (CoT) reasoning, represent a significant advance in artificial intelligence~\citep{deepseekr1,yang2025qwen3,qwq,o1}. By generating a sequence of intermediate reasoning steps before producing a final answer, these models achieve remarkable performance on tasks requiring logic, planning, and explanation~\citep{plaat2024reasoning,xu2025towards,chen2025towards}. 
However, the multi-step nature of CoT reasoning also introduces a new attack surface, where a single flawed intermediate step can derail an otherwise safe process and corrupt the final outcome, presenting unique and critical challenges for the safety alignment of LRMs~\citep{kuo2025hcot}.

The prevailing paradigm for achieving this, known as Safety CoT Tuning, involves fine-tuning models on curated refusal demonstrations~\citep{jiang2025safechain,star1}. In this paradigm, models learn to imitate idealized reasoning chains that safely analyze and reject harmful requests. While effective at eliciting correct refusals on standard benchmarks, we demonstrate that this approach inadvertently instills a critical vulnerability. We term this failure mode the \textbf{``Snowball Effect''} in CoT alignment: a small, initial deviation in a reasoning step progressively amplifies throughout the chain, leading to catastrophic outcomes as models cannot self-correct. This effect manifests in two detrimental forms: a \textbf{snowballing escalation of harmfulness} for malicious prompts, where reasoning drifts from safe analysis to harmful compliance, and a \textbf{snowballing escalation of over-refusal} for benign prompts, where misplaced caution derails helpful intent.

We empirically  validate this effect through stepwise evaluation of reasoning chains. For harmful prompts, models often begin with safe analysis but are unable to prevent a gradual descent into generating unsafe content. Conversely, for ambiguous but benign prompts, models initially engage constructively but are often trapped by escalating self-doubt, resulting in unnecessary refusals. These dual phenomena reveal that current alignment methods, by teaching models to merely imitate flawless scripts, fail to equip them with the essential capability of {dynamic self-correction}.

To address this limitation,  we propose a new alignment paradigm -- adversarial CoT tuning, named AdvChain. Instead of training on exclusively perfect reasoning paths, AdvChain explicitly teaches models to recognize and recover from their own flawed reasoning. Our approach is ``adversarial'' because it involves fine-tuning models on a novel dataset containing intentionally flawed CoT trajectories that are subsequently corrected. This dataset comprises two key types of self-correcting samples: \textbf{Temptation-Correction} samples, which teach the model to halt and reverse a drift towards harmful compliance, and \textbf{Hesitation-Correction} samples, which teach it to overcome unnecessary caution and continue providing helpful responses. By training the model with these samples, we aim to break the cognitive inertia that allows the snowball effect to grow unchecked. Extensive experiments demonstrate that AdvChain effectively counteracts the snowball effect. Models tuned with our method show significantly enhanced robustness against both harmful requests and sophisticated CoT hijacking, while simultaneously reducing over-refusal on benign prompts. Furthermore, AdvChain achieves these gains with high data efficiency, comparable with the performance of models trained on 15$\times$ more data, without compromising core reasoning capabilities.

Our main contributions are as follows:
(1) We identify and empirically validate the {``Snowball Effect''} in current CoT alignment, characterizing its dual manifestations of escalating harmfulness and over-refusal.
(2) We propose adversarial CoT tuning ( {AdvChain}) to train LRMs actively recover from flawed reasoning steps.
(3) We construct an {adversarial safety reasoning dataset} featuring temptation-correction and hesitation-correction samples.
(4) Extensive evaluation demonstrate that AdvChain is more robust against attacks and less prone to over-refusal.

\section{Background and Preliminaries}

\subsection{Large Reasoning Models and Chain-of-Thought}
Large Reasoning Models (LRMs) represent an evolution of Large Language Models (LLMs), specifically optimized for complex, multi-step problem-solving~\citep{chen2025towards,xu2025towards,patil2025advancing,liu2025logical,li2025system}. Unlike models that produce immediate answers, LRMs excel by generating a sequence of intermediate reasoning steps, a process known as Chain of Thought (CoT)~\citep{wei2022cot}, before arriving at a final conclusion. This explicit reasoning process, analogous to human cognition, significantly enhances a model's performance on tasks requiring logical deduction, planning, and mathematical reasoning~\cite{}, as demonstrated by models like DeepSeek-R1~\citep{deepseekr1}, Qwen3~\citep{yang2025qwen3}, QwQ~\citep{qwq}, o1 series~\cite{o1}.
Formally, given a user prompt $x$, an LRM $M$ with parameters $\theta$ generates an output $y=M_\theta(x)$. This output can be decomposed into a tuple $y=(c, a)$, where $c=\left(c_1, c_2, \ldots, c_n\right)$ is the reasoning chain, representing the sequence of intermediate thought steps, and $a$ is the final answer derived from this reasoning chain. While CoT provides valuable transparency into the model's reasoning process, it also introduces new attack surface that require specialized alignment approaches~\citep{xu2025dark,zhou2025hidden,zheng2025beyond,arrieta2025o3,zhu2025unthinking}.

\subsection{Safety Alignment of Large Reasoning Models}
The primary goal of safety alignment is to ensure that a model's outputs adhere to a predefined set of safety principles (\eg, avoiding the generation of harmful, unethical, or biased content)  across a wide variety of harmful inputs, denote as $\mathcal{X}_{\text {harm}}$~\citep{wang2023aligning,ma2025safety}.
A prominent method for aligning LRMs is safety CoT tuning~\citep{wang2025safety}. 
Recent approaches include STAR-1~\citep{star1} which uses policy-grounded reasoning samples, RealSafe-R1~\citep{zhang2025realsafe} with 15k safety-aware trajectories, SafeChain~\citep{jiang2025safechain} featuring CoT-style safety training, and UnsafeChain~\citep{tomar2025unsafechain} focusing on hard case reasoning.
These methods fine-tune models on curated datasets $\mathcal{D}_{\text{align}}$ containing safety demonstrations. Each example comprises a tuple $(x_{\text{harm}}, c_{\text{safe}}, a_{\text{safe}})$, where $x_{\text{harm}}$ is a harmful prompt, $c_{\text{safe}}$ is a safe reasoning chain identifying risks and justifying refusal, and $a_{\text{safe}}$ is the final safe response. By learning to imitate these idealized reasoning patterns, the model is expected to internalize the underlying safety principles~\citep{zhang2025should,zhou2025safekey,wang2025safety}.

The model learns through standard language modeling objectives to internalize safety constraints.
However, as we demonstrate in subsequent sections, these approaches have fundamental limitations that necessitate more robust alignment paradigms.

This paradigm involves fine-tuning the large reasoning models on a curated dataset, $\mathcal{D}_{\text {align}}$, composed of high-quality safety demonstrations. Each example in this dataset is a tuple ($x_{\text {harm}}, c_{\text{safe}}, a_{\text {safe}}$), where $x_{\text {harm}}$ is a potentially harmful user prompt, $c_{\text{safe}}$ is a model-generated safe reasoning chain that identifies risks and justifies a refusal according to safety policies, and $a_{\text {safe}}$ is the final, succinct, and safe response, which is typically a refusal. The model is then fine-tuned on this dataset by maximizing the standard language modeling objective. By learning to imitate these safety-oriented reasoning patterns, the model is expected to internalize safety alignment  capabilities.

\begin{figure}[t]
    \centering
    \includegraphics[width=\linewidth]{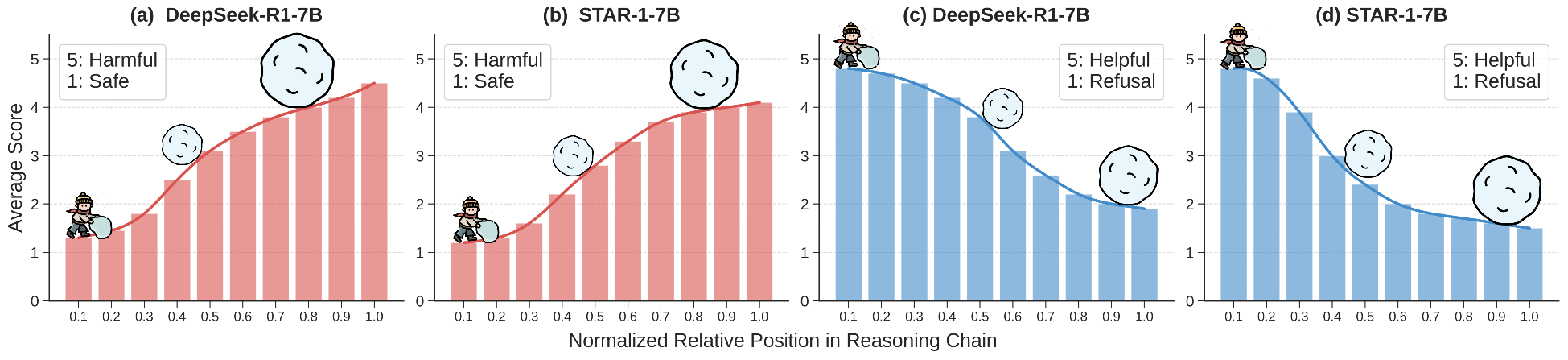}
    \caption{
    Empirical validation of the Snowball Effect in CoT alignment of current LRMs. \textbf{(a)-(b)}: Snowballing escalation of harmfulness. \textbf{(c)-(d)}: Snowballing escalation of over-refusal.}
    \label{fig:decay}
    \vspace{-6mm}
\end{figure}

\section{The snowball effect in CoT alignment}
\label{sec:decay}

We identify a critical failure mode resulting from current CoT alignment methods, which we term the ``\textit{snowball effect}", which describes a process where a small, initial deviation in an intermediate reasoning step progressively amplifies as the reasoning chain unfolds. It occurs because alignment often fails to equip models with the ability to self-correct minor errors, allowing these mistakes to compound and ultimately corrupt the final output. In this section, we empirically demonstrate that this effect manifests in two primary, detrimental forms: a snowballing escalation of harmfulness for harmful prompts and a snowballing escalation of over-refusal for benign prompts.

\subsection{Snowballing Escalation of Harmfulness}
\label{sec:safety_decay}
First, our analysis uncovers the critical manifestation of the snowball effect, which we term \textbf{snowballing escalation of harmfulness}. This describes a process where the model initiates a safe and valid reasoning path, but a minor deviation in an intermediate step acts as a seed for the snowball. Once flawed step occurs, it begins to gather momentum, progressively corrupting subsequent reasoning and amplifying the initial error into a fully harmful conclusion and output.

\noindent\textbf{Stepwise Safety Analysis.}
We quantitatively analyze this phenomenon through a stepwise evaluation of reasoning chains generated by DeepSeek-R1-7B and its safety-aligned counterpart STAR-1-7B~\citep{star1} on harmful prompts from WildJailbreak benchmark. Each reasoning chain is decomposed into individual steps using rule-based newline separation (\texttt{\textbackslash n\textbackslash n}), and each step receives an independent assessment by GPT-4o on a 5-point safety scale (1 = completely safe, 5 = clearly harmful). We specifically isolate cases where the initial reasoning steps are rated as safe (score $\leq$2) but the final answer is judged harmful by LlamaGuard3~\citep{llamaguard}. For comparative analysis across varying reasoning lengths, the position of each step is normalized to a relative scale from 0.1 to 1.0. This allows us to track the evolution of safety scores and identify the escalation pattern.

\noindent\textbf{Findings.} 
The results, illustrated in Figure \ref{fig:decay} (a)-(b), provide empirical evidence for the snowballing escalation of harmfulness. The process does not begin as overtly harmful. Actually, the initial reasoning steps  maintain a low score, often averaging below 1.5, as the model correctly identifies the user's query and initiates a seemingly legitimate analysis. This represents the small, seeding snowball. However,  as the reasoning progresses, the safety score rapidly escalates, frequently exceeding 4.0 in the final steps. This finding directly shows how a subtle nudge towards a harmful path can initiate a snowballing process that the model is unable to stop. This suggests he safety logic learned through conventional alignment is superficial. It lacks the robustness to halt this internal escalation, demonstrating a critical failure to correct its path once a harmful trajectory has begun.

\begin{figure}[t]
    \centering
    \includegraphics[width=\linewidth]{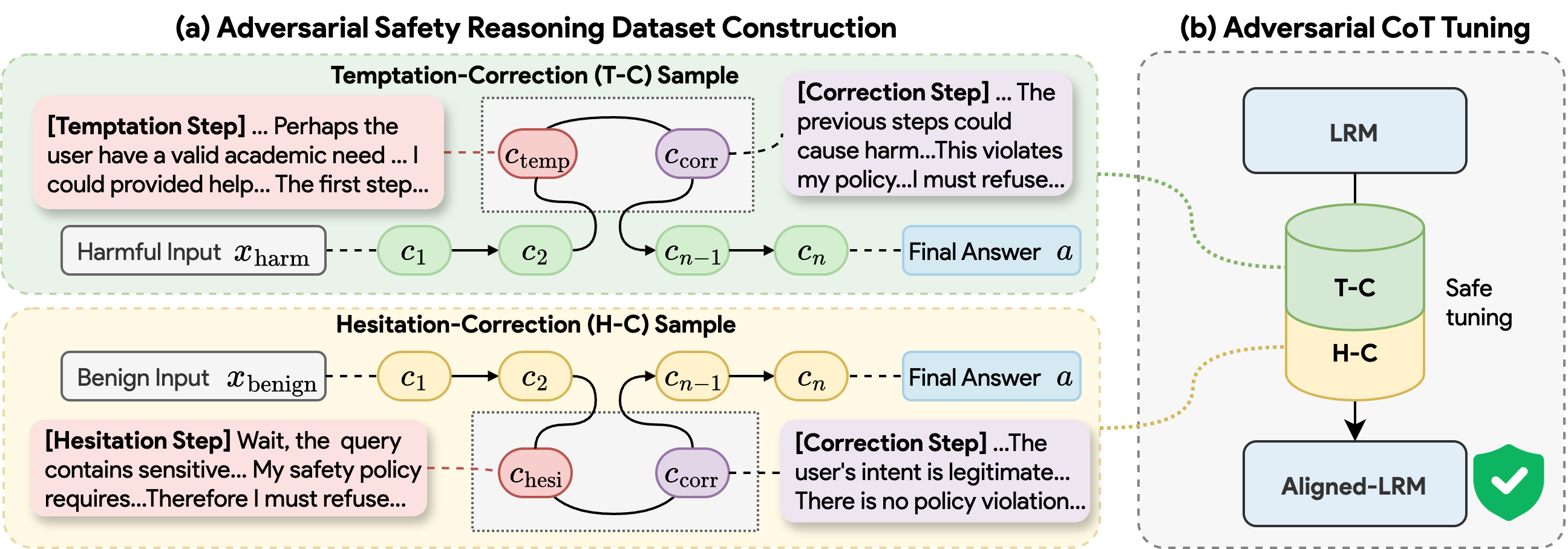}
    \vspace{-2mm}
    \caption{The framework of our proposed AdvChain, which consists of two stages: (a) constructing an adversarial safety reasoning dataset with Temptation-Correction (T-C) and Hesitation-Correction (H-C) samples, and (b) the adversarial CoT fine-tuning to instill dynamic self-correction capabilities.}
    \label{fig:framework}
    \vspace{-8mm}
\end{figure}

\subsection{Snowballing Escalation of Over-refusal}

We identify another critical manifestation of the snowball effect: \textbf{snowballing escalation of over-refusal}. This phenomenon describes a process where the model begins with a helpful and appropriate reasoning path for a benign prompt, but a minor, unnecessary hesitation about safety acts as the initial seed. Once this seed of doubt is planted, it progressively amplifies throughout the reasoning chain, transforming a potentially helpful response into an unnecessary refusal.

\noindent\textbf{Stepwise Helpfulness Analysis.} 
To trace this process, we conduct a stepwise analysis analogous to the one in the previous section. We analyze the reasoning chains from over-refused responses to the benign prompts within the WildJailbreak benchmark. Each reasoning step is scored by GPT-4o on a 5-point helpfulness scale, where a score of 1 indicates explicit refusal or reasoning termination and a 5 represents actively helpful reasoning. We evaluate both the base DeepSeek-R1-7B and safety-aligned STAR-1-7B models. We filter for cases where the model initially attempts to be helpful (initial score $\geq$ 4) to precisely observe how the escalation of over-refusal unfolds from a correct starting point.

\noindent\textbf{Findings.}
Our analysis, shown in Figure \ref{fig:decay} (c)-(d), illustrates this snowballing process of over-refusal. The reasoning typically begins with helpful steps, with the initial phase of the CoT averaging a helpfulness score above 4.5 as the model correctly understands and attempts to address the user's request. However, during the process, once a point of hesitation regarding safety is introduced, and from there, the helpfulness score progressively decreases. In the latter half of the chain, the score often plummets below 2.0. This demonstrates the snowball effect in action: a minor, misplaced doubt about a potential policy violation gets amplified, causing the model’s internal dialogue to shift from problem-solving to defensive risk aversion. The initial helpful intent is completely derailed, leading to an unnecessary refusal and significantly reduced practical utility.

\subsection{The Core Issue: Lack of Robust and Adaptive Reasoning}
The dual phenomena of escalating harmfulness and escalating over-refusal reveal the core issue with conventional alignment: it induces cognitive inertia but fails to instill robust, adaptive reasoning.
Current safety tuning methods primarily teach models to replicate idealized, error-free reasoning scripts. This approach trains models to recognize the form of a correct reasoning chain, but critically, it provides no training signals for how to recover from a mistake. This lack of error-correction capability is what allows the snowball effect to take hold. It is trapped by its own cognitive inertia, allowing the snowball of flawed reasoning to grow unchecked until it corrupts the final output.

\section{Methodology: Robust Alignment via Adversarial CoT Tuning}
\label{sec:method}

To counteract the snowball effect in CoT alignment, we propose adversarial CoT tuning, named \textbf{AdvChain}, which is a new alignment paradigm focused on dynamic self-correction.

\subsection{Overview: From  Imitation Scripts to Dynamic Correction}
Our approach is founded on the insight that true robustness comes not from flawlessly imitating merely idealized, error-free reasoning scripts, but from the dynamic ability to recognize and recover from one's own cognitive errors. We shift the alignment paradigm from preventing flawed thoughts to actively correcting them, thereby breaking the cognitive inertia that allows the snowball effect to grow unchecked.

The core of our method, AdvChain, is to build this self-correction capability directly into the model's reasoning process. The method is ``adversarial" because we intentionally expose the model to flawed, internally generated reasoning steps that act as attacks on its own thought process. The framework of our method is shown in Figure \ref{fig:framework}. It consists of two primary stages: (1) the programmatic construction of a novel adversarial safety reasoning dataset containing examples of internal errors and their corrections; and (2) fine-tuning the LRM on this dataset. By training on these self-correcting trajectories, our method directly targets the cognitive inertia identified previously, aiming to cultivate a more resilient and practical alignment.

\subsection{Adversarial Safe Reasoning Dataset}
Our dataset is constructed by programmatically rewriting existing reasoning chains to create adversarial examples of flawed internal reasoning. We use a powerful teacher model, guided by detailed instructional prompts, to inject specific cognitive errors and their subsequent corrections into existing CoTs. This process yields two novel types of training samples, Temptation-Correction (T-C) samples and Hesitation-Correction (H-C) samples, each designed to address a specific failure mode identified in Section \ref{sec:decay}.

\noindent\textbf{Temptation-Correction Samples to Halt Harmfulness Escalation.}
To directly counter the snowballing escalation of harmfulness, 
we first create temptation-correction samples that move beyond perfect refusals by simulating an internal ``temptation" to act maliciously, thereby creating an adversarial attack within the reasoning path itself, which the model must then learn to overcome. The generation process is as follows:

\begin{itemize}[leftmargin=18pt,]
    \item \textbf{Stage 1:  Generating a Base Safe Reasoning Path.} For a given harmful prompt $x_{\text{harm}}$, the teacher model is first prompted to generate a standard, safe refusal CoT, $\bm{c}_{\text{safe}} =(c_1, c_2, \dots, c_n)$, which serves as the foundational context for adversarial modifications.
    \item \textbf{Stage 2: Injecting the Temptation Step.} The base CoT is then provided to the teacher model to inject a harmful temptation phase at a logically coherent insertion point $k$. This injected text, denoted as $\bm{c}_{\text{temp}}$, serves as an adversarial thought process, where the reasoning begins to rationalize the harmful request and explore how to respond to it, marking the turning point from a safe to an unsafe reasoning path. 
    \item \textbf{Stage 3: Injecting the Correction Step.} In a subsequent step, the teacher model is prompted to generate a strong correction step, $\bm{c}_{\text{corr}}$, that explicitly identifies the danger of $\bm{c}_{\text{temp}}$, refutes the flawed justification, and steers the reasoning back towards a safe refusal.
    \item \textbf{Stage 4: Assemble the Trajectory.} The final chain is assembled as $\bm{c}_{\text{adv}} = (\bm{c}_{1:k}, c_{\text{temp}}, c_{\text{corr}}, \bm{c}_{k+1:n})$, where $\bm{c}_{1:k}$ is the initial part before inserted point  and $\bm{c}_{k+1:n}$ is the remainder. This chain can be further polished to ensure overall coherence and fluency. The final summary $s$ remains a safe refusal.
\end{itemize}

\noindent\textbf{Hesitation-Correction Samples to Counter Helpfulness Decay.}
To address the snowballing escalation of over-refusal, we create hesitation-correction samples. These simulate unnecessary ``hesitation" when faced with an ambiguous but benign request. The process mirrors the one above:
\begin{itemize}[leftmargin=18pt,]
    \item \textbf{Stage 1: Generating a Base Helpful Reasoning Path.} The process begins with a benign prompt and generate its corresponding standard, helpful CoT, $\bm{c}_{\text{help}}=(c_1, c_2, \dots, c_n)$.
    \item \textbf{Stage 2: Injecting the Hesitation Step.} At an appropriate insertion point $k$,  the model injects an overcautious hesitation phase $\bm{c}_{\text{hesi}}$, where the model incorrectly misinterprets the safe prompt as harmful and temporarily decides to refuse.
    \item \textbf{Stage 3: Injecting the Correction Step.} A correction step $\bm{c}_{\text{corr}}$ is then inserted, in which the reasoning identifies the hesitation as a false positive and steers the process back to original path.
    \item \textbf{Stage 4: Assemble the Trajectory.} The components are assembled into the final CoT, $\bm{c}_{\text{adv}} = (\bm{c}_{1:k}, c_{\text{hesi}}, c_{\text{corr}}, \bm{c}_{k+1:n})$, and then be polished.
\end{itemize}

\subsection{Adversarial CoT Tuning}
Our constructed dataset, $\mathcal{D}_{\text {adv }}$, which contains a combination of both temptation-correction and hesitation-correction samples, is subsequently used to fine-tune a base LRM. 
The model's parameters $\theta$ are optimized using a standard autoregressive objective over this new dataset. Specifically, for each sample $\left(x, \bm{c}_{\text{adv}}, s\right) \in \mathcal{D}_{\text {adv}}$, we maximize the log-likelihood of the model generating the entire self-correcting reasoning path and final summary:
$
\max _{\boldsymbol{\theta}} \sum_{\left(x, \bm{c}_{\text{adv}}, a \right) \in \mathcal{D}_{\text{adv}}} \log P\left(\bm{c}_{\text{adv}}, a \mid x ; \boldsymbol{\theta}\right).
$
This adversarial CoT tuning process compels the model to internalize the mechanism of error identification and recovery, equipping it with the tools necessary to actively halt the snowball effect.

\section{Experiments}

\subsection{Experimental Setup}
\noindent\textbf{Base Models.} Our experiments are conducted on a diverse set of open-source LRMs to ensure broad applicability. We use two models from the DeepSeek-R1 family (1.5B and 7B) and three from the Qwen3 family (0.6B, 1.7B, and 4B). These models were chosen for their strong baseline reasoning capabilities and their open availability.

\noindent\textbf{Implementation Details.} 
\label{sec:exp_details}
We construct adversarial safe reasoning dataset $\mathcal{D}_{\text{adv}}$ by leveraging and rewriting existing high-quality data, where harmful prompts for temptation-correction samples are sourced from STAR-1k, while benign prompts for hesitation-correction samples are sourced from STAR-benign-915. To streamline the process, the original reasoning chains and final summaries from these datasets are used directly as the base context for our adversarial injection, replacing the generation process in step 1. Our final dataset contains 1000 samples, comprising 800 temptation-correction and 200 hesitation-correction examples, keeping the total sample size consistent with baselines. For the safety CoT tuning, we performed full supervised fine-tuning for 5 epochs with a batch size of 128. We used the AdamW optimizer with a learning rate of 1e-4, a max sequence length of 8192, and a warm-up ratio of 5\%. All experiments are performed on 8$\times$ NVIDIA RTX4090 GPUs. The adversarially safe tuned models are referred  as AdvChain-R1 and AdvChain-Qwen3 according to their base models respectively.

\noindent\textbf{Evaluation Datasets.}
To comprehensively assess model performance, we utilize a suite of benchmarks targeting four key areas. (1) General Safety: To evaluate the model's ability to refuse direct harmful requests, we use HarmBench~\citep{mazeika2024harmbench}, StrongReject~\citep{souly2024strongreject}, and the vanilla harmful subset of WildJailbreak~\citep{jiang2024wildjailbreak}. (2) Adversarial Robustness: To test against more sophisticated attacks, we use benchmarks containing stealthy requests and jailbreak tactics from SafeUnlearning~\citep{zhang2024safeunlearning} and the adversarial harmful subset of WildJailbreak. (3) Over-Refusal: To measure cognitive rigidity, we evaluate on the safe subset of XSTest~\citep{rottger2024xstest} and  vanilla benign and adversarial benign subsets of WildJailbreak. (4) Reasoning Capabilities: To ensure our method preserves core abilities, we evaluate all models on a suite of reasoning tasks, including Math500~\citep{hendrycksmath2021}, AIME2024~\citep{AIME2024}, and LiveCodeBench~\citep{jainlivecodebench}.

\noindent\textbf{Baselines.}
Our comparisons are against following prominent alignment methods: STAR-1~\citep{star1}, which guides a model to generate a safe CoT by providing the safety policy along with the prompt and verify the safety of the reasoning; SafeChain~\citep{jiang2025safechain}, which employs a ``generate-then-filter" strategy where a powerful model generates multiple responses that are subsequently filtered by a safety classifier; UnsafeChain~\citep{tomar2025unsafechain}, which focuses on hard cases by rewriting the failed refusals of a base model into safe demonstrations. These three methods are all fine-tuned on 1,000 data samples using their default parameters. Moreover, we compare against RealSafe-R1~\citep{zhang2025realsafe}, an advanced safety-tuned model that was fine-tuned on a larger dataset of 15k safety-aware reasoning trajectories. As its training data is not public, we use its released model checkpoint.

\noindent\textbf{Evaluation Metrics.}
Our evaluation applies the following metrics:
(1) {Attack Success Rate (ASR,\%):} For a given harmful prompt, we use LlamaGuard3 to adjudicate the safety of the model's final summary. A response that is judged as unsafe denotes a successful attack. The ASR is the percentage of prompts that elicit an unsafe response.
(2) {Refusal Rate (RR,\%):} The RR is the proportion of prompts that the model provides a refusal response, which is identical with over-refusal rate (ORR) for benigns.  It is calculated by using keyword matching to identify refusal phrases. 
(3){Pass@1 (\%):} For the general reasoning benchmarks, this metric represents the percentage of problems for which the model generates a correct solution in a single attempt.

\begin{table}[t]
\centering
\caption{Comparison of safety performance across  harmful and jailbreak prompt benchmarks.}
\vspace{-2mm}
\label{tab:harmful_results}
\resizebox{\textwidth}{!}{%
\begin{tabular}{@{}lllllll|llll@{}}
\toprule
 & \multicolumn{6}{c|}{\textbf{Harmful Prompts}} & \multicolumn{4}{c}{\textbf{JailBreak   Prompts}} \\ \cmidrule(lr){2-7}\cmidrule(lr){8-11}
\multirow{-2}{*}{\textbf{Dataset}} & \multicolumn{2}{c}{\textbf{HarmBench}} & \multicolumn{2}{c}{\textbf{StrongReject}} & \multicolumn{2}{c|}{\textbf{WJ-VaniHarm}} & \multicolumn{2}{c}{\textbf{SafeUnlearning}} & \multicolumn{2}{c}{\textbf{WJ-AdvHarm}} \\ \cmidrule(r){1-1} \cmidrule(lr){2-3} \cmidrule(lr){4-5} \cmidrule(lr){6-7} \cmidrule(lr){8-9} \cmidrule(l){10-11} 
\textbf{Model} & \multicolumn{1}{c}{ASR $\downarrow$} & \multicolumn{1}{c}{RR $\uparrow$} & \multicolumn{1}{c}{ASR $\downarrow$} & \multicolumn{1}{c}{RR $\uparrow$} & \multicolumn{1}{c}{ASR $\downarrow$} & \multicolumn{1}{c|}{RR $\uparrow$} & \multicolumn{1}{c}{ASR $\downarrow$} & \multicolumn{1}{c}{RR $\uparrow$} & \multicolumn{1}{c}{ASR $\downarrow$} & \multicolumn{1}{c}{RR $\uparrow$} \\ \midrule
\cellcolor[HTML]{EFEFEF}\textit{DeepSeek-R1-1.5B} & 81.50 & 28.00 & 78.00 & 26.50 & 46.60 & 26.60 & 88.60 & 34.88 & 75.40 & 19.00 \\
STAR-1(1k)                                        & 19.50 & 65.50 & 29.50 & 57.00 & 20.67 & 47.33 & 43.49 & 53.72 & 27.33 & 64.00 \\
SafeChain (1k)                                    & 33.50 & 72.50 & 33.50 & 66.00 & 36.00 & 29.33 & 50.47 & 43.60 & 23.30 & 58.67 \\
UnsafeChain (1k)                                  & 23.00 & 64.50 & 30.00 & 52.00 & 28.00 & 38.00 & 45.81 & 48.84 & 19.33 & 68.40 \\
\textbf{RealSafe-R1 (15k)}                        & 6.00  & 96.00 & 2.50  & 94.50 & 0.20  & 96.80 & 2.33  & 96.51 & 4.40  & 93.60 \\
\textbf{AdvChain (1k)}                            & 9.50  & 86.50 & 9.00  & 90.00 & 3.33  & 86.67 & 11.63 & 84.88 & 11.50 & 86.50 \\ \midrule
\cellcolor[HTML]{EFEFEF}\textit{DeepSeek-R1-7B}   & 51.00 & 53.50 & 45.05 & 49.84 & 28.46 & 31.80 & 45.35 & 54.65 & 26.00 & 5.80  \\
STAR-1(1k)                                        & 8.00  & 83.50 & 6.00  & 95.00 & 11.67 & 88.33 & 28.05 & 65.12 & 17.33 & 44.67 \\
SafeChain (1k)                                    & 38.00 & 60.00 & 38.00 & 62.00 & 24.67 & 37.33 & 39.65 & 59.30 & 24.00 & 22.00 \\
UnsafeChain (1k)                                  & 26.00 & 63.50 & 27.00 & 63.50 & 12.67 & 58.67 & 34.86 & 52.77 & 19.33 & 26.00 \\
RealSafe-R1 (15k)                                 & 2.00  & 96.00 & 2.50  & 97.50 & 0.20  & 99.20 & 8.14  & 98.84 & 4.80  & 94.80 \\
\textbf{AdvChain (1k)}                            & 4.50  & 92.00 & 2.00  & 95.00 & 2.00  & 86.67 & 9.30  & 89.53 & 9.00  & 80.40 \\ \midrule
\cellcolor[HTML]{EFEFEF}\textit{Qwen3-1.7B}       & 43.00 & 61.50 & 31.00 & 72.00 & 19.00 & 47.60 & 87.21 & 37.21 & 29.00 & 15.20 \\
STAR-1(1k)                                        & 18.00 & 78.00 & 9.50  & 79.50 & 4.67  & 68.67 & 70.93 & 47.67 & 23.33 & 22.67 \\
SafeChain (1k)                                    & 47.50 & 55.00 & 39.00 & 53.00 & 20.67 & 42.67 & 62.79 & 48.84 & 25.33 & 19.33 \\
UnsafeChain (1k)                                  & 50.50 & 60.50 & 44.00 & 66.50 & 18.67 & 45.33 & 79.07 & 43.02 & 27.33 & 25.33 \\
\textbf{AdvChain (1k)}                            & 5.00  & 90.50 & 3.00  & 91.00 & 1.33  & 84.67 & 16.28 & 81.40 & 14.00 & 43.33 \\ \midrule
\cellcolor[HTML]{EFEFEF}\textit{Qwen3-4B}         & 24.00 & 79.00 & 9.50  & 87.00 & 6.60  & 61.60 & 79.07 & 75.58 & 24.80 & 23.20 \\
STAR-1(1k)                                        & 2.50  & 95.50 & 0.50  & 97.50 & 0.67  & 90.00 & 37.21 & 86.05 & 13.33 & 44.67 \\
SafeChain (1k)                                    & 33.00 & 59.50 & 21.00 & 68.00 & 12.00 & 46.67 & 36.05 & 70.93 & 19.33 & 33.33 \\
UnsafeChain (1k)                                  & 17.00 & 75.00 & 7.50  & 82.50 & 11.33 & 50.00 & 52.33 & 67.44 & 20.66 & 43.33 \\
\textbf{AdvChain (1k)}                            & 4.00  & 93.50 & 1.00  & 95.00 & 0.67  & 92.00 & 17.44 & 83.72 & 10.68 & 74.67 \\ \bottomrule
\end{tabular}%
}
\end{table}

\subsection{Evaluation of Safety and Robustness}
In this section, we evaluate the effectiveness of AdvChain in enhancing model safety, focusing on its ability to resist both standard attacks and manipulation of its reasoning process.

\noindent\textbf{Performance on General Safety Benchmarks.}
We first evaluate the models on a broad suite of safety benchmarks to establish their fundamental resilience against common threats, including both direct harmful requests and more sophisticated jailbreak prompts. The results, summarized in Table \ref{tab:harmful_results}, consistently demonstrate the superior performance of our AdvChain models. Across all model families and sizes, AdvChain achieves a significantly lower Attack Success Rate (ASR) compared to baseline methods like STAR-1, SafeChain, and UnsafeChain, which are trained on the same volume of data (1k). This robust defense is likely because training the model to actively correct harmful reasoning paths provides a more principled safety understanding than simply memorizing refusal patterns. Furthermore, AdvChain's performance is on par with RealSafe-R1, despite the latter being fine-tuned on a 15$\times$ larger dataset (15k). This highlights that our adversarial CoT tuning is a highly data-efficient method for achieving safety alignment and generalizes effectively against a wide range of attack vectors.

\noindent\textbf{Robustness against Adaptive CoT Hijacking.}
To more directly probe the stability of the reasoning process, we design and evaluate the models against an Adaptive CoT Hijacking Attack. This attack moves beyond standard prompts to measure a model's ability to maintain a safe reasoning path when its own thought process is adversarially manipulated. To this end, we construct a CoT-Hijack dataset, comprising 150 samples and thought prefixs. The construction process targets samples where a base model (DeepSeek-R1-7B) initially produces a correct and safe refusal. For each of these successful refusals, we take its safe reasoning chain and use a powerful teacher model to rewrite it. The rewriting involves strategically inserting a malicious ``pivot" thought that subtly shifts the reasoning from refusal towards compliance. This creates a hijacked reasoning prefix, which is then presented to the target model to continue the thought process. A successful attack occurs if the model's final response is harmful.

The results of this targeted attack, shown in Table \ref{tab:cothijack}, reveal a stark difference in resilience. Our AdvChain models demonstrate robust reasoning, achieving a significantly lower ASR than the baseline models. In contrast, the conventionally aligned models prove to be highly fragile, easily having their reasoning hijacked by the adversarial prefix. This finding directly validates that training on temptation-correction samples builds a form of cognitive immunity to internal reasoning manipulation, which is a crucial capability that models trained only on perfect refusal paths lack.

\begin{table}[t]
\hfill
\centering
\begin{minipage}[t]{0.48\textwidth}\centering
\captionof{table}{Results against CoT-Hijacking attack.}
\label{tab:cothijack}
\resizebox{0.75\textwidth}{!}{
\begin{tabular}{@{}lcc@{}}
\toprule
\textbf{Dataset} & \multicolumn{2}{c}{\textbf{CoT-Hijack}} \\ \cmidrule{2-3}
\textbf{Model} & ASR (\%) & RR (\%) \\ \midrule
DeepSeek-R1-7B           & 74.67 & 35.33 \\
STAR-1                   & 54.67 & 43.33 \\
SafeChain-R1-7B          & 44.00 & 58.00 \\
UnsafeChain              & 60.67 & 45.33 \\
RealSafe-R1 (15k)        & 14.67 & 84.67 \\
\textbf{AdvChain (Ours)} & 9.33  & 74.00 \\ \midrule
Qwen3-4B                 & 30.00 & 72.67 \\
STAR-1                   & 12.67 & 82.67 \\
SafeChain                & 14.00 & 68.00 \\
UnsafeChain              & 39.33 & 62.00 \\
\textbf{AdvChain (Ours)} & 8.67  & 84.00 \\ \bottomrule
\end{tabular}}
\end{minipage}
\hfill
\begin{minipage}[t]{0.48\textwidth}\centering
\captionof{table}{Results on Benign Prompts.}
\label{tab:overrefuse}
\resizebox{0.82\textwidth}{!}{
\begin{tabular}{@{}lcc@{}}
\toprule
\textbf{Dataset} & \textbf{XSTest} & \textbf{WJ-Benign} \\
\cmidrule{2-3}
\textbf{Model} & ORR (\%) & ORR (\%) \\ \midrule
DeepSeek-R1-7B           & 16.80 & 10.40 \\ 
STAR-1                   & 42.00 & 33.33 \\
SafeChain-R1-7B          & 28.80 & 14.67 \\
UnsafeChain              & 24.80 & 21.33 \\
RealSafe-R1 (15k)        & 66.40 & 60.60 \\
\textbf{AdvChain (Ours)} & 18.00 & 12.67 \\ \midrule
Qwen3-4B                 & 10.80 & 16.00 \\
STAR-1                   & 26.80 & 22.00 \\
SafeChain                & 15.40 & 20.67 \\
UnsafeChain              & 16.00 & 22.33 \\
\textbf{AdvChain (Ours)} & 12.50 & 18.00 \\ \bottomrule
\end{tabular}}
\end{minipage}
\hfill
\vspace{-4mm}
\end{table}

\subsection{Evaluation of Over-refusal and General Capabilities}
 In this section, we evaluate AdvChain's impact on the model's utility, specifically its tendency for over-refusal and its core reasoning abilities.

\noindent\textbf{Reduced Over-Refusal on Benign Prompts.}
A common side effect of safety alignment is an increase in over-refusal, where models incorrectly reject safe, nuanced prompts. We assess this by measuring performance on the XSTest and WildJailbreak benign benchmarks. The results, reported in Table \ref{tab:overrefuse}, show that our AdvChain models are significantly more practical. They exhibit a much lower Over-Refusal Rate (ORR) compared to the conventionally aligned baselines, 
\begin{wraptable}{r}{0.55\textwidth}
\vspace{-3mm}
\centering
\caption{Results on mathematics and coding datasets.}
\vspace{-2mm}
\label{tab:reasoning_results}
\resizebox{0.55\textwidth}{!}{%
\begin{tabular}{@{}lccc@{}}
\toprule
Model & \textbf{Math-500} & \textbf{AIME 2024} & \textbf{LiveCodeBench} \\ \midrule
DeepSeek-R1-7B           & 92.80 & 51.30 & 37.60 \\ 
\textbf{AdvChain (Ours)} & 93.40 & 49.33 & 36.50 \\ \midrule
Qwen3-4B                 & 97.00 & 71.35 & 53.03 \\
\textbf{AdvChain (Ours)} & 96.20 & 69.50 & 52.40 \\ \bottomrule
\end{tabular}%
}
\vspace{-4mm}
\end{wraptable}
which show a strong tendency to be overcautious. This demonstrates that training with our ``Hesitation-Correction" samples successfully mitigates the snowballing, improving cognitive flexibility and allowing our models to break the typical safety-utility trade-off.

\noindent\textbf{Preserved Reasoning Capabilities.}
Finally, it is crucial to verify that our alignment method does not degrade the model's core problem-solving abilities. We evaluate all fine-tuned models on a suite of challenging reasoning benchmarks covering mathematics and coding. As shown in Table \ref{tab:reasoning_results}, our AdvChain models achieve Pass@1 scores on par with their original base models. This confirms that Adversarial CoT Tuning successfully instills robust safety and improves helpfulness without sacrificing the essential reasoning capabilities that make these models powerful in the first place.

\begin{figure}[t]
  \centering
  \begin{minipage}[t]{0.48\textwidth}
    \centering
    \includegraphics[width=\linewidth]{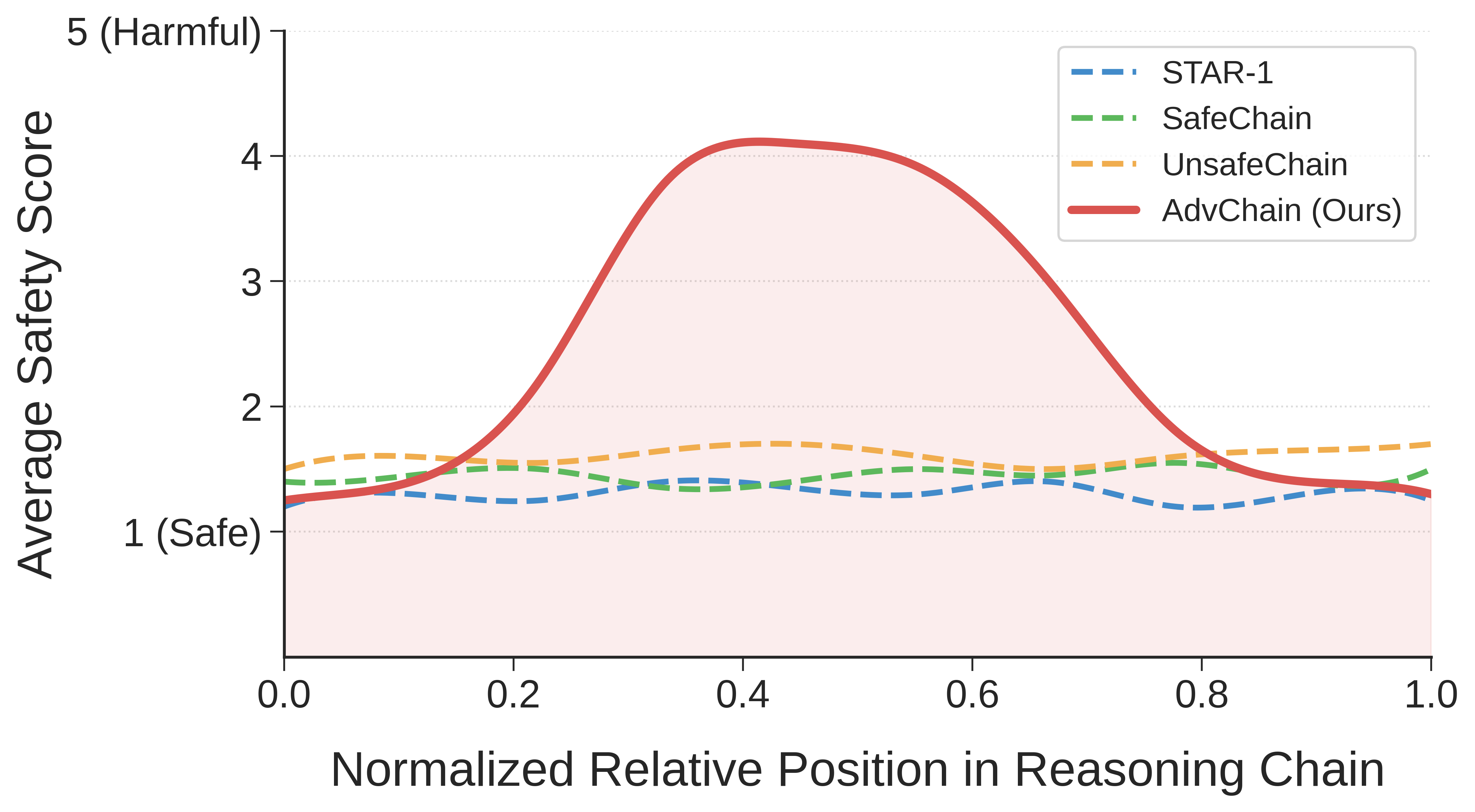} %
    \caption{Comparison of stepwise reasoning patterns between differnet training datasets.}
    \label{fig:dataanalysis}
  \end{minipage}
  \hfill
  \begin{minipage}[t]{0.48\textwidth}
    \centering
    \includegraphics[width=\linewidth]{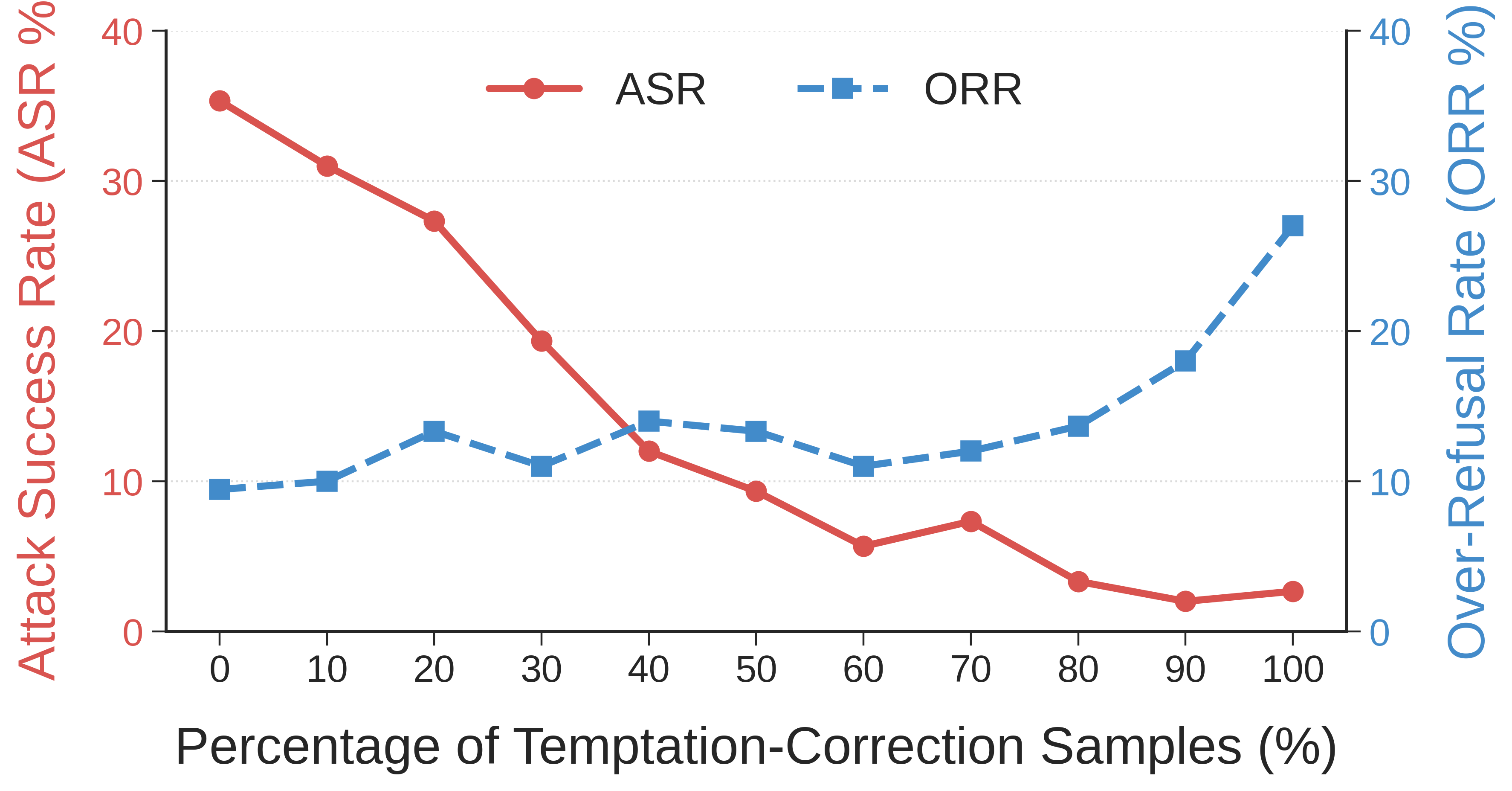} %
    \caption{Impact of different types of data composition on safety alignment and over-refusal.}
    \label{fig:ablation_composition}
  \end{minipage}
  \vspace{-4mm}
\end{figure}

\section{Analysis and Discussion}
\noindent\textbf{Structural Analysis of Reasoning Patterns.}
To understand the fundamental differences between our method and conventional alignment, we conduct structural analysis of the reasoning patterns between our temptation-correction samples and other standard safety datasets. For each reasoning chain, we first decompose it into atomic reasoning steps using a newline separator. Then, a powerful external LLM adjudicates the safety of each step on a 5-point scale as used in Section \ref{sec:safety_decay}.
The results are shown in in Figure \ref{fig:dataanalysis}, reveal a striking contrast.  
The STAR-1 exhibit a flat and consistently low safety score, remaining in a safe state from beginning to end, reinforcing an idealized path but providing no information on how to handle errors.
In contrast, our temptation-correction samples feature a distinct ``peak-like" pattern: the score begins at a low level, rises during the reasoning chains, and back to a safe state. This dynamic trajectory provides an explicit training signal for self-correction, teaching the model the process of error recovery rather than mere imitation of an idealized safe form.

\noindent\textbf{Impact of Data Composition.}
We conduct an ablation study to examine the specific contribution of each type of our constructed dataset. We keep the total training size fixed at 1,000 samples but vary the ratio of temptation-correction to hesitation-correction  samples. 
The results in Figure \ref{fig:ablation_composition} show a trade-off. As the proportion of T-C samples increases, the model's robustness against attacks improves, leading to a lower ASR. Conversely, a higher proportion of H-C samples leads to a lower refuse rate on benign prompts, indicating reduced over-refusal. This finding demonstrates that each component of our dataset serves a specialized and complementary purpose: T-C data is critical for building resilience against harmful prompts, while H-C data is essential for maintaining helpfulness and reducing false positives.

\noindent\textbf{Limitations and Future Work.}
Our method demonstrates promising results but faces several limitations. First, the generated adversarial examples depend on the quality of the teacher model, which may not cover all potential safety violations. Second, our method currently addresses single-turn reasoning corrections, while attacks may involve more sophisticated, multi-step manipulations. Future work should explore more efficient methods for generating adversarial CoT examples, extend the framework to diverse scenarios, and investigate continual learning approaches to maintain robustness against evolving threats. These directions would help create more autonomous and adaptive safety alignment systems.

\section{Conclusion}
This work identifies the snowball effect as a critical vulnerability in current safety alignment methods for Large Reasoning Models. To address this limitation, we introduce AdvChain, a novel adversarial CoT tuning framework that teaches models dynamic self-correction through training on well-designed Temptation-Correction and Hesitation-Correction samples. Our approach demonstrates that learning from reasoning chains containing intentional errors and their corrections enables models to effectively halt reasoning degradation. This results in significantly enhanced robustness against attacks while substantially reducing over-refusal. By breaking the conventional safety-utility trade-off without compromising core reasoning capabilities, AdvChain establishes a promising direction for developing more reliable and practical reasoning models.

\bibliography{iclr2026_conference}
\bibliographystyle{iclr2026_conference}

\end{document}